\documentclass{article}
\usepackage{ijcai20}

\usepackage{times}

\usepackage{soul}
\usepackage{url}
\usepackage[hidelinks]{hyperref}
\usepackage[utf8]{inputenc}
\usepackage[small]{caption}
\usepackage{graphicx}
\usepackage{amsmath,amssymb}
\usepackage{booktabs}
\def\by{{\mathbf y}}

\def\bh{{\mathbf h}}

\def\bx{{\mathbf x}}
\def\bz{{\mathbf z}}
\def\bw{{\mathbf w}}
\def\bW{{\mathbf W}}
\def\bX{{\mathbf X}}
\def\bF{{\mathbf F}}
\def\bZ{{\mathbf Z}}
\def\bY{{\mathbf Y}}

\def\bB{{\mathbf B}}
\def\bR{{\mathbf R}}
\def\bG{{\mathbf G}}
\def\bS{{\mathbf S}}
\def\bH{{\mathbf H}}

\def\bM{{\mathbf M}}
\def\bL{{\mathbf L}}
\def\bI{{\mathbf I}}
\def\bz{{\mathbf z}}

\def\bg{{\mathbf g}}
\def\R{{\mathbb R}}

\def\balpha{\mbox{{\boldmath $\alpha$}}}

\def\b0{{\mathbf 0}}
\def\b1{{\mathbf 1}}

\newtheorem{definition}{Definition}[section]
\title{Heterogeneous Representation Learning: A Review}

\author{
Joey Tianyi Zhou$^1$\and %\footnote{Contact Author}
Xi Peng$^2$\And
Yew-Soon Ong$^{3}$\\
\affiliations
$^1$IHPC, A*STAR, Singapore\\
$^2$College of Computer Science, Sichuan University, China\\
$^3$Nanyang Technology University, Singapore \\
\emails
zhouty@ihpc.a-star.edu.sg, pengx.gm@gmail.com, asysong@ntu.edu.sg
}

\begin{document}

\maketitle

\begin{abstract}
The real-world data usually exhibits heterogeneous properties such as modalities, views, or resources, which brings some  unique challenges wherein the key is Heterogeneous Representation Learning (HRL) termed in this paper.   
This brief survey covers the topic of HRL, centered around several major learning settings and real-world applications. First of all, from the mathematical perspective, we present a unified learning framework which is able to model most existing learning settings with the heterogeneous inputs. 
After that, we conduct a comprehensive discussion on the HRL framework by reviewing some selected learning problems along with the mathematics perspectives, including multi-view learning, heterogeneous transfer learning, Learning using privileged information and heterogeneous multi-task learning.  For each learning task, we also discuss some applications under these learning problems and instantiates the terms in the mathematical framework. Finally, we highlight the challenges  that are less-touched in HRL and present future
research directions. To the best of our knowledge, there is no such framework to unify these heterogeneous problems, and this survey would benefit the community.
\end{abstract}

%%%%%%%%%%%%%%%%%
\section{Introduction}
In a lot of real-world problems, such as social computing, video surveillance, and healthcare, data are collected from diverse domains or obtained from various feature extractors/sensors, thus leading to the heterogeneous properties in terms of modalities (e.g., texts, images and videos) or views (e.g., multi-lingual and cross-sensors). For example, the posts on social media like Facebook usually simultaneously employ texts, images and voice to describe different profiles of the same topics/events. In other words, the cross-modal or multi-view data does not exist in an isolation fashion. Instead, it is usually the combination of various data types in a many-to-one correspondence fashion because the complete information of a given object/event distributes in multiple views/modalities. 

\begin{figure}[!t]
  \vspace{-0.2cm}
\centering
  \includegraphics[width=0.8\columnwidth]{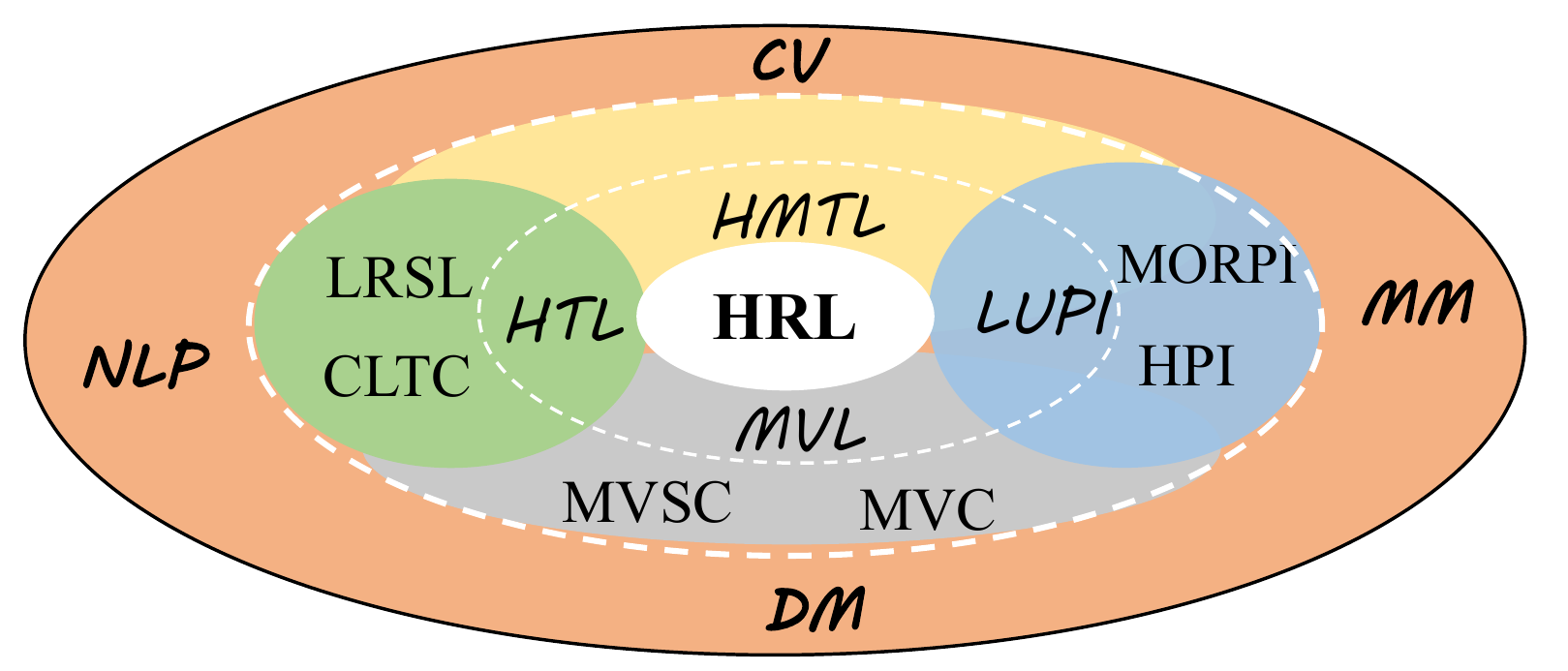}
  \vspace{-0.2cm}
  \caption{\small{Relationship between HRL and learning tasks/applications.}
  \scriptsize{``MVL": Multi-View Learning, ``HTL": Heterogeneous Transfer Learning, ``HMTL": Heterogeneous Multi-task Learning, ``LUPI": Learning Using Privileged Information; ``MVSC": Multi-view Subspace clustering, ``MVC": Multi-view Classification, ``CLTC": Cross-lingual Text Classification, ``LRSL": Low-Resource Sequence Labeling, ``MORPI": Multi-object Recognition with PI, ``HPI“: Hashing with PI.
  }}
  \label{fig:intro}
  \vspace{-0.5cm}
\end{figure}

Besides the many-to-one correspondence, another heterogeneity lies on the one-to-one correspondence. One typical example is cross-lingual machine translation wherein all data are in text but in different languages and all bi-lingual data pairs contain the ``same and complete'' information of the object/event of interest. Moreover, there exist a lot of tasks of such a paradigm, e.g., person re-identification, tracking, 3D data point matching. 
% More specifically, video frames are captured by multiple cameras for the same person. Although they are in the video modality, the spatial-temporal context significantly varies across cameras and the task aims to establish. 

In the aforementioned problems, it is far away from extracting desirable features with the same or homogeneous descriptors due to the heterogeneity of data. In addition to the heterogeneity of data, the heterogeneity of task is another challenge, which refers to that different tasks may have different inputs, outputs, and the associated downstream tasks. It is still unclear how to discover the hidden useful information contained in multiple related heterogeneous tasks to improve the generalization performance of all the tasks. 

To address these challenges, there are a variety of techniques have been proposed including but not limited to  multi-view/-modal learning, heterogeneous transfer learning, heterogeneous multitask learning, etc. Unfortunately, these  techniques are mostly isolated studied in each domain. 
To the best of our knowledge, there is no study to unify these topics and give comprehensive review from the perspective of Heterogeneous Representation Learning (HRL). We believe that HRL could be an effective unified solution to trigger recent explosion of interest in all these vibrant multi-disciplinary fields.

In this review, we first give a formal definition and narrow down our research focus to differentiate it from other related works but in different lines. 
\begin{definition}
Given a given data set $\{\mathcal{D}_k\}_{k=1}^{K} \!=\! \{(\bx_{k_i})_{i=1}^{n_k}\}_{k=1}^{K}$ drawn from $K$ domains and the cross-domain correspondences (e.g., one-to-one or many-to-one),  the original feature spaces for each domain  $\bx_{k_i} \in \R^{d_k}$ are heterogeneous (e.g., $d_i \neq d_j, \forall i \neq j$ ), Heterogeneous  Representation Learning aims at learning feature/task mappings  $\{f_k(\cdot)\}_{k=1}^K$ which connect different domains to facilitate the downstream tasks $\{\mathcal{T}_k\}_{k=1}^{K}$.
% A set $S\in \mathbb R^n$ is said affine set if for every $x,y \in S$ and every $\lambda \in \mathbb R$ we have $\lambda x+(1-\lambda )y\in S$.
\end{definition}

To be noted that, the aforementioned correspondences are shared across domains, which is necessary to HRL learning in different manners. For example, many-to-one correspondences could be the labels accompanying the data.  If  the data from two domains are annotated with the same label, they would exhibit some similar properties in the feature space. In addition, to learn a common representation shared by different views/modalities, one-to-one correspondence is always explicitly taken in multi-view analysis wherein the data are assigned in sample-wise. It is clear that how to utilize these correspondences becomes a crucial research problem in HRL. 

Unlike existing surveys which are either from the perspective of a general learning setting like  transfer learning \cite{pan2009survey},  multi-view learning~\cite{xu2013survey},  multi-task learning~\cite{zhang2017survey}, or from the perspective of downstream applications~\cite{kowsari2019text,wang2017survey}, we provide a revisit of our recent efforts in several selected learning tasks and their downstream applications from the view of HRL (see Fig. \ref{fig:intro}). To the best of our knowledge, this could be also the first study on discussing the diverse learning settings and applications in a unified perspective of HRL based on  mathematical formulation.
Beyond their similarity, we further discuss the differences as well as the research goals along HRL.

\section{Mathematical View}
In this section, we review the study on the problems by providing a mathematical view of HRL. Specifically, all the tasks related to HRL could be described in the following  framework that includes three major terms in the objective function $\mathcal{J}$, i.e., 1) a within-domain loss, 2) an inter-domain loss, and 3) a task-related regularization. 
\small{
\begin{align}
\label{eq:HRL}
\min&\sum_{k}\underbrace{L(\{\mathcal{D}_k\};\Theta_L)}_{\text{Within-Domain Loss}}+ \lambda_1 \underbrace{M(\{\mathcal{D}_k\};\Theta_M)}_{\text{Inter-Domain Loss}}+ \lambda_2 \underbrace{R(\{\mathcal{D}_k\};\Theta_R)}_{\text{Regularization}}\nonumber\\
\text{s.t.} ~~&\Theta_L \in \Omega_L, \Theta_M\in \Omega_M, \Theta_R\in \Omega_R, k=\{1,2,\cdots,K\},
\end{align}}
where $\{\mathcal{D}_k\}_{k=1}^K$ denotes the input data.  $\Theta_L$,$\Theta_M$,and $\Theta_R$ denote the learnable parameters of the within-domain loss $L(\cdot)$, 
inter-domain loss $M(\cdot)$, and task-related regularization $R(\cdot)$, respectively. For different tasks, their learnable parameter sets may be further constrained by $\Omega_L$, $\Omega_M$, and $\Omega_R$. Noticed, the ``domain'' has different definitions in different tasks and applications. For example, it refers to ``view" in multi-view analysis. 

The within-domain loss $L(\cdot)$ is a domain-related function used to describe the tasks/property within each domain. The inter-domain loss $M(\cdot)$ aims to model the relationship across domains under the help of the cross-domain correspondences which could be constructed or given in different scenarios. For example, in multi-view learning, the one-to-one correspondence based on views are usually given in data collection or preparation. In transfer learning, the  multi-class label could be deemed as many-to-one correspondences. In multi-label learning, one-to-many correspondences could be established between the data point and multiple labels. The third term in Eq.(\ref{eq:HRL}) is the regularization term to avoid over-fitting or achieve data prior, which is specifically defined in different tasks and architectures, for example, sparsity and low rankness. 

In the following sections, we are going to discuss our recent efforts on multi-view learning, heterogeneous transfer learning, learning using privileged information, and heterogeneous multi-task learning in the proposed framework. \footnote{To be noted that we are not giving a  review of all the related learning tasks and applications, which is beyond the scope of this brief review. Instead, we summarize and introduce some of our works in the aforementioned unified framework to verify the homogeneity of these tasks/applications.}

\section{Multi-view Learning}
As one of most important topics in machine learning,  multi-view learning (MVL) aims at fusing the knowledge
from multiple views to facilitate the down-streaming learning tasks, e.g., clustering, classification, and retrieval. The key challenge of MVL is exploring the data correspondence across multiple views. The mappings among different views are able to couple the view-specific
knowledge, while the additional task-related regularization  is incorporated based on different priors of data latent structure. 
% In the following, we will discuss some representative applications in multi-view learning in more details. 

% \subsection{Cross-modal Retrieval}
% The existing multi-modal retrieval could be mainly group into supervised and unsupervised learning based on the availability of label information.  For
% the unsupervised methods,  only  co-occurrence information can be utilized to learn common representations across multi-modal data. The co-occurrence information means that if different
% modalities of data are co-existed in a multimodal document,
% then they are of the similar semantic. For example, the textual
% description along with images or videos often exist in a
% webpage to illustrate the same event or topic. Furthermore, the
% unsupervised methods seek joint representation for multimodal data

% As known to all, canonical correlation analysis (CCA)\cite{} is
% the most popular algorithm to achieve a common space for
% two views. Specifically, CCA attempted to obtain two projections, one for each view, to transform the data from two different views into a shared subspace, respectively, through
% maximizing the cross correlation across two views:
% \begin{equation}
% M(\cdot) = -corr (h_A(X_{A},
% \Theta_A), h_B(X_{B}, \Theta_B)),
% \end{equation}
% where corr is matrix correlation operation.  The mappings $h_A$  and $h_B$ are parametrized by $\Theta_A$ and $\Theta_B$. In CCA, mappings is linear. In KCCA, mappings are represented by kernels. In DCCA, mappings are implemented with the neural network. 

\subsection{Multi-view Subspace clustering}
Multi-view subspace clustering
(MVSC) targets at grouping similar objects into the
same subspace and dissimilar objects into different subspaces by exploring the available multi-view information. The core commonality of existing MVSC methods is encapsulating the
complementary information of different views to learn a
shared/common representation followed by a single-view
clustering approach. More specifically, MVSC first learns a latent representation to bridge the gap of multiple view and then applies the traditional clustering methods like
spectral clustering algorithms or k-means on the representation to obtain the final clustering partitions. 

Formally, let $\mathcal{D}_k = \{\bx^{(k)}_i\}_{i=1}^{N}$ denote the data from $k$-th view and each view consists of $N$ observation, one aims to infer a shared latent representation
$\bh$ for each data point across views.  To obtain the latent representation, the objective of most existing methods could be generally decomposed into two major kinds of loss, i.e., view-specific loss $L(\cdot)$, and inter-domain consistency loss $M(\cdot)$, namely,
\begin{equation}
\begin{small}
\mathcal{J} =\underbrace{L(\{\mathcal{D}_k\}_{k=1}^K)}_{\text{View-specific loss}} 
+ \lambda_1\underbrace{M(\{\mathcal{D}_k\}_{k=1}^K)}_{\text{Cross-view Loss}}+ \lambda_2\underbrace{R(\cdot)}_{\text{Regularizer}}.\nonumber
\end{small}
\end{equation}
%  The first group of methods are based on the canonical correlation analysis (CCA)\cite{chaudhuri2009multi}  and its variants which usually construct the cross-view loss with the following mathematical formulation, 
% \begin{equation}
% M(\cdot) = -\sum_i^N corr(h_1(\bx_i^{(1)}),h_2(\bx_i^{(2)}))
% \end{equation}
% Specifically, CCA attempted to obtain two projections $h_m(\bx) = \bw_m^\prime\bx$, one for each view, to transform the data from two different views into a shared subspace, respectively, through maximizing the correlation across two views. In CCA, mappings are linear. In KCCA\cite{fukumizu2007statistical}, mappings are represented by kernels, namely $h_m(\bx) = \balpha_m^\prime k_1(\bx,\cdot)$. In DCCA\cite{andrew2013deep}, mappings are implemented with the neural network  wit $d$ layers, with $h_m(\bx) = s(W_d^m z_{d-1} +b_d^m)$, where $z$ is hidden layer, $W$ is a matrix of weights, $b$ is a vector of biases and $s()$ is a nonlinear function applied component-wise.

Different methods define different loss functions. %For example, canonical correlation analysis (CCA)\cite{chaudhuri2009multi}  and its variants \cite{fukumizu2007statistical,andrew2013deep} which usually construct the cross-view loss $M(\cdot) = -\sum_i^N corr(h_1(\bx_i^{(1)}),h_2(\bx_i^{(2)}))$ through learning view-specific  projection that transform the data from two different views into a shared subspace.
For example, \citeauthor{ijcai2019-356} propose a multiview spectral clustering network  with the following formulations, $L (\cdot) = (1-\lambda_1) \sum_k \frac{1}{N^2}\sum_{i,j}^N \bW_{ij}^{(k)}\|\bh_i^{(k)}-\bh_j^{(k)}\|$, where $\bW^{(k)}$ denotes a precomputed similarity graph for the $k$-th view. $M (\cdot) = \lambda \frac{1}{NK^2}\sum_{k,p,i}\|\bh_i^{(k)}-\bh_j^{(p)}\|$, where $\bh_i^{(k)}$ denotes the output of a parametric model w.r.t. the input $\bx_i^{(k)}$. More specifically, $\bh_i^{(k)} = f^{(k)}(\bx_i^{(k)})$, where $f^{(k)}$  denotes the $k$-th learnable subnetwork in \cite{ijcai2019-356} or the linear mapping used in \cite{li2015large}. 
\citeauthor{kangNN} propose a novel objective which integrates the objectives of graph construction and spectral clustering. Specifically,  $L(\cdot) = \sum_k\|\bX^{(k)}-\bX^{(k)}\bZ^{(k)}\|_2^2 + \lambda_1 Tr(\bF_k^\prime \bL^{(k)}\bF_k)+ \lambda_2 w_k\|\bY\bY^\prime-\bF_k\bF_k^\prime\|$, where $\bF_k$ is the partition result, $\bY$  is the consensus cluster indicator matrix, and $\bL^{(k)}$ denotes the graph Laplacian matrix derived from $\bZ^{(k)}$. In addition, $\bX^{(k)}$ and $\bZ^{(k)}$ denote feature matrix and subspace representation of the $k$-th view, and $\R(\cdot) = \|\bZ^{(k)}\|_F^2$ is used to avoid overfitting. 

% There are two main categories for recent Multi-view Subspace
% Clustering (MSC) methods. One category conducts self representation within each view  and simultaneously explores correlations among different views. Let $\bX^{(k)}$
% an $\bZ^{(k)}$ denote feature matrix and subspace representation corresponding to the $k$-th view, respectively, then we obtain the following general formulation: 
% \begin{equation}
% \mathcal{J} = L(\{\bX^{(k)}\}^K_{k=1};\{\bZ^{(k)}\}^K_{k=1}) + \lambda R(\{\bZ^{(k)}\}^K_{k=1}) 
% \end{equation}

More recently,  \citeauthor{peng2019comic} propose to cluster multi-view data through automatic parameters learning. Specifically, the method defines the 
view-specific loss to be a combination of view-specific reconstruction loss and geometric consistency, i.e., $L(\cdot)  =  \frac{1}{2}\|\bx_i^{(k)}-\bz_i^{(k)}\|_2^2+ +\frac{1}{2}\lambda\sum_{i,j}\bW_{ij}^{(k)}\left(\|\bS_{ij}^{(k)}\bz_i^{(k)}-\bS_{ij}^{(k)}\bz_j^{(k)}\|_2^2+\mu(\bS_{ij}^{(k)}-1)^2 \right)$,
where $\bz_i^{(k)}$ denotes the learned representation of $i$-th data point from the $k$-th view and $\bS^{(k)}$ is connection graph of view-$k$. 
$M(\cdot) = \frac{1}{2}\sum_{i,j}(\bS_{ij}^{(A)}-\bS_{ij}^{(B)}),$
where $M(\cdot)$ defines cross-view cluster assignment consistency.

Besides, some subspace clustering methods share the following formulation,
\begin{equation}
\mathcal{J} = L(\{\bX^{(k)}\}^K_{k=1};\bH) + \lambda_1 M(\bH;\bZ) +  \lambda_2 R(\bZ). \nonumber
\end{equation}
For example, \citeauthor{Li_2019_ICCV} define a view-specific loss $L(\cdot) = \frac{1}{2}\sum_k\|\bX^{(k)}-\bX^{(k)}\bZ^{(k)}\|_2^2 + \frac{1}{2}\sum_k\|\bZ^{(k)}-g_{\Theta_E^k}(\bH)\|_2^2$, where the first part applies the reconstruction loss to alleviate the noise effect. The second part learns a common latent representation $\bH$ by using it to reconstruct all views-specific representations, where $\Theta_E^k$ is the weighting parameter for network $g$.  $M(\bH;\bZ) = \frac{1}{2}\|\bH-\bH\bZ_C\|_F^2$, where the common subspace representation $\bZ_C$ reveals the subspace structure of latent representation $\bH$.  $R(\cdot) = \sum_k(\|\bZ^{(k)}\|_*+ \|\Theta_E^k\|_F^2 ) + \|\bZ_C\|_*$,  where the nuclear norm is used to guarantee the high within-class homogeneity. 

% In latent multi-view subspace clustering methods \cite{zhang2018generalized,zhou2019dual}, the view-specific loss $L$ and cross-view loss $M$  have the following formulations respectively,  $L(\cdot) = L_h(\{\bX^{(v)}\}^V_{k=1};\{\bP^{(v)}\}^K_{k=1}\bH)$, $ M(\cdot) = L_r(\bH,\bH\bZ)$, where $L_h$ denotes the loss functions associated with latent representation and $L_r$  defines based on the self-representation-based reconstruction error on the latent space. In \cite{zhang2018generalized}, $L_h(\{\bX^{(v)}\}^V_{v=1};\{\bP^{(v)}\}^V_{v=1}\bH) = \sum_v\|\bX^{(v)}-\bP^{(v)}\bH\|_{2,1}$ and $L_r(\{\bH,\bH\bZ) = \|\bH-\bH\bZ\|_{2,1}$, and   $R(\bZ) = \|\bZ\|_*$  is an additional term that regularizes reconstruction coefficient matrix, e.g., $\|Z\|_{21}$, nuclear norm $\Z\|_*$. In \cite{zhou2019dual}, the shared subspace is further divided into one shared and view-specific representations. Thus the low-dimension representation for view is defined as  $[\bH;\bH_v]$, where $\bH$ and $\bH_c$ denote shared and view-specific component. Accordingly, the self-representation for view is defined as $[\bZ;\bZ_v]$. 

\subsection{Multi-view Classification}
The multi-view classification (MVC) aims to build a classification model with the given multiple views of training data. In the
test stage, we also consider all the available views of each testing instance to make the final prediction. Different from multi-view subspace clustering, multi-view classification usually only involves one step rather than two-step optimization by utilizing the supervised information $\bY$ to guide representation learning:
\begin{equation}
\begin{small}
\mathcal{J} =\underbrace{L(\{\mathcal{D}_k\}_{k=1}^K)}_{\text{View-specific loss}} 
+ \lambda_1\underbrace{M(\{\mathcal{D}_k\}_{k=1}^K)}_{\text{Cross-view Loss}}+ \lambda_2\underbrace{R(\{\mathcal{D}_k\}_{k=1}^K, \bY)}_{\text{Discriminative Loss}}\nonumber
\end{small}
\end{equation}

To regularize the representation learning, most methods usually employ a single quotient to incorporate the supervised information across views. As a result, their main difference lies on the choice on the cross-view loss which could be generalized as 
$
M(\cdot) = f(\frac{S_W}{S_B}),
$
where $S_W, S_B$ denotes the within-class compactness and the between-class scatter that are computed across all views.  
For example,  \citeauthor{kim2006learning} extend canonical correlation analysis (CCA)\cite{chaudhuri2009multi}   to Discriminative CCAs wherein the within-class correlation is maximized
and  the between-class correlation is simultaneously minimized
in the learned common space. By adopting the fisher discriminant analysis, \citeauthor{diethe2008multiview} extend this idea by explicitly using the label information. %In \cite{kan2015multi}, additional cross-view loss named view-consistency is further added by resembling the transforms of different views. \cite{kan2016multi} constructs the view-specific and common subnetworks with MLP which is regularized by fisher loss. In \cite{lin2006inter}  proposed to  minimize the intra-class scatter and meanwhile maximize the inter-class separability.

In addition, another group of methods incorporate the supervised information into the classification rather than representation learning loss~\footnote{As suggested in~\cite{Peng2020:TNNLSa}, classification could also be treated as a special case of representation learning.}. For example, 
%Coupled Spectral Regression (CSR) \cite{lei2009coupled} implemented Classification loss $R(\cdot)$  by learning a projection from the observation to the common low-dimensional embedding of the class label through least squares regression. The view-specific loss is defined as $L(\cdot) = \sum \bW_{ij}^{(k)} \|\bh_{i}^{(k)}-\bh_{j}^{(k)}\|$, where $\bW$ is deployed by  graph embedding to preserve the  geometric properties of the data set and
% $\bh_{i}^{(k)}$ denotes the learned representation of  $i$-th datapoint from $k$-th domain. 
\citeauthor{NIPS2019_8346} propose the reconstruction loss $L(\cdot) = \sum W_{i}^{m} \|f(\bh_{i}^{(k)})-\bx_{i}^{(k)})\|$ which encodes all partial views into the shared representation $\bh$, where  $W_{i}^{k}$ indicates the view availability for $i$-th data point. $R(\cdot) = \ell_c(y_i, y,\bh_i)$ is a clustering-like loss which learns the structured representation in a common space.  \citeauthor{Xiao2018ActionRF} use the cross-entropy loss for multi-view action recognition, $R(\cdot) = \sum_i^N y_i\log \hat{y}_i $, where $\hat{y}_i$ is predicted labels from the network. 

In positive-unlabeled learning task, \citeauthor{pmlr-v25-zhou12} proposes a co-regularization framework to construct the cross-view loss $L(\cdot) = \|r^{(1)} - r^{(2)}\|$ so that the density ratio functions $r^{(1)}(\bx^{(1)})$ and  $r^{(2)}(\bx^{(2)})$ make agreement on the same instance. The classification loss  
is defined to be the unconstrained least-square importance fitting, i.e., $R(\cdot) = \frac{1}{2}\int \left( r(\bx)-\frac{p(s=1,\bx)}{p(\bx)}\right)p(\bx)d\bx$, where $r(\bx)$ is density ratio between the positive samples and all the samples and $s$ denotes indicator of the positive annotation.

\section{Heterogeneous Transfer Learning}
Heterogeneous Transfer Learning (HTL) targets at improving the performance on the target domain $\mathcal{D}_T = \{\bx_{T_i},y_{T_i}\}_{i=1}^{n_T}$  by exploiting auxiliary labeled source domain data $\mathcal{D}_S = \{\bx_{S_i},y_{S_i}\}_{i=1}^{n_S}$ ($n_S \gg n_T$), when $\mathcal{D}_S$ shares the different feature space as the target domain $\mathcal{D}_T$.  To bridge two the source and target domain, most existing HTL methods either require cross-domain correspondences or shared labels to learn mappings and most of them could be summarized into the following formulation, 
 \begin{equation}
%\begin{split}
\mathcal{J} =\underbrace{L(\{\mathcal{D}_S,\mathcal{D}_T\})}_{\text{Domain-specific Loss}}+ \lambda_1 \underbrace{M(\{\mathcal{D}_S,\mathcal{D}_T\})}_{\text{Domain Discrepancy loss}} + \lambda_2\underbrace{R(\cdot)}_{\text{Regularizer}}.\nonumber
%\end{split}
\end{equation}

\subsection{Cross-lingual Text Classification}
Cross-lingual text classification (CLTC) refers to the task of classifying documents in different languages into the same taxonomy of categories.  Text classification largely relies on manually annotated training data, which however is expensive to create training data. Therefore, it is promising to explore how to use the training data given in only one source language (i.e, source domain) to classify text
written in a different target language (i.e, target domain) with no or a small number of labeled data. 
Most existing studies minimize such feature differences between data in the two languages and conduct translation through learning the transformations based on the either labels or one-to-one correspondences. 

Most existing CLTC methods could be roughly categorized into two groups. The first group usually assumes there are a number of cross-lingual correspondences such as cross-lingual translation pairs, $\{\bx^{(c)}_{S_i},\bx^{(c)}_{T_i}\}_{i=1}^{n_C}$.
With the correspondences, some works borrow the auto-encoder loss as the domain discrepancy loss $M(\cdot)$ via 
\begin{equation}
 M(\{\mathcal{D}_S,\mathcal{D}_T\}) = \sum_i^{n_C}\|\bh_{S_i}^{(c)}-f(\bh_{T_i}^{(c)})\|^2,\nonumber
\end{equation}
where $\bh_{S_i}^{(c)}$ and $\bh_{T_i}^{(c)}$ are the hidden  representations of the source and target domain. More specifically, \citeauthor{DBLP:conf/aaai/ZhouPTY14,ijcai2017-454} formulate the cross-domain mapping $f(\cdot)$ using a linear mapping, i.e., $M(\{\mathcal{D}_S,\mathcal{D}_T\}) = \sum_i^{n_C}\|\bh_{S_i}^{(c)}-\bW \bh_{T_i}^{(c)}\|^2$, where the representation $\bh$ is learned from linear reconstruction objective \cite{DBLP:conf/icdm/ShiLFYZ10} or marginalized stack denoising autoencoder \cite{DBLP:conf/aaai/ZhouPTY14}.  \citeauthor{DBLP:conf/aaai/ZhouPTY14} define the loss on the source  domain $L(\{\mathcal{D}_S\})$ by the stack denoising autoencoder loss, namely,  $\sum_i^{m}\|\overline{\bX}_{S}-\bW_S \overline{\bX}^{(i)}_{S}\|^2$, where  $\overline{\bX}_{S}$ denotes the union source domain data matrix, and $\overline{\bX}^{(i)}_{S}$ denotes the $i$-th corrupted version.
\citeauthor{DBLP:journals/ai/ZhouPT19} further extend the linear cross-domain mapping into a nonlinear mappings using the multilayer perception. Formally, $M(\{\mathcal{D}_S,\mathcal{D}_T\}) = \sum_i^{n_C}\|\bh_{S_i}^{(c)}- \bW^{(2)}_T \tanh(\bW^{(1)}_T \bh_{T_i}^{(c)})\|^2$. 
 
%In \cite{xiao2013novel}, the $H_{A,k}^{(c)}$ are learnt  through matrix completion. The matching loss is realized by  sparsity and low rank matrix regularization. On the top of it, 
Another group of methods projects the heterogeneous features space into a higher-dimension space \cite{DBLP:conf/aaai/ZhouPTH16} so that  the features from two domains are homogeneous via the techniques like Maximum Mean Discrepancy (MMD), i.e., $M(\mathcal{D}_S,\mathcal{D}_T) = \|\frac{1}{n_S}\sum_i\phi(\bx_{S_i})-\frac{1}{n_T}\sum_j\phi(\bx_{T_j})\|_\mathcal{H}^2$. 
% \cite{DBLP:conf/ijcai/Yan0WMTW18} use the entropic Gromov-Wasserstein discrepancy to define the $M(\{\mathcal{D}_S,\mathcal{D}_T\}) = \epsilon_{M_T,M_T}(\bT) - \gamma H(\bT)$, where $\bT$ is transportation matrix, $H(\bT)$ is the entropy of $\bT$ and metric matrices are $\bM_S = \bX_S^\top\bX_S$ and $\bM_T =\bX_T^\top\bX_T$.

% Different from previous methods, Zhou~\etal~\cite{DBLP:conf/aistats/ZhouT14,JMLR:v20:13-580} proposed to learn a sparse feature mapping between the source domain and the target domain by exploiting commonality between multiple binary classification tasks decomposed from the target multi-class classification problem. 
% \begin{equation}
%  M(\cdot) =\sum_{t}^{n} \|\bw_{T}^t-\bG\bw_S^t\|^2 + \lambda \sum_i^{d_T}\|\bg_i\|_1,
% \end{equation}
% where the linear mapping $\bG$ is to minimize  the difference between $\bw_{T}^t$ and $\bG\bw_S^t$ over all the $n_c$ tasks, the
% second term in the objective is to enforce the sparsity on each row of $\bG$.

The other branch of methods just assume that only a few annotated target data points are given and they construct the domain classification loss by directly exploiting the label information of two domains. Therefore, they deploy the loss functions on two domains as below,
\begin{equation}
%\begin{small}
L(\{\mathcal{D}_S,\mathcal{D}_T\}) = \sum_{i}^{n_S}L(\{\mathcal{D}_S\};\Theta_S) + \sum_{j}^{n_T} L(\{\mathcal{D}_T\};\Theta_T)\nonumber
%\end{small}
\end{equation}
where $L(\cdot)$ could be realized with the parameters of source and target domain, denoted by $\Theta_S,\Theta_T$. 
For example, \citeauthor{xiao2014feature} deploy the squared loss $L(\{\mathcal{D}_S\};\cdot) = \|\by_S-f(\bx_S)\|^2$ and $L(\{\mathcal{D}_T\};\cdot) = \|\by_T-\bM f(\bx_T)\|^2$, where $f$ is a prediction function represented by kernel matrix and $\bM$ denotes the kernel matching matrix.  
% \cite{DBLP:conf/ijcai/WangM11,DBLP:conf/cvpr/KulisSD11,ijcai2017-349} proposed  to align heterogeneous features in a latent space based on dis-/similarity constraints constructed from label information in both the source and  target domain. 
% In \cite{ijcai2017-349} the domain classification loss is defined to be hinge rank loss $L(\{\mathcal{D}_S\};\Theta_S) = \frac{1}{n_S}\sum_i^{n_S}\sum_{y\neq y_{S_i}}\max[0,\epsilon -f(g(x_{S_i}))\cdot(y_{S_i}-y)]$,  where $g_{\Theta_S}, g_{\Theta_T}$ denotes the source and target domain transformation mappings respectively, and $f$ denote shared transformation onto the embedded label space.
\citeauthor{DBLP:conf/icml/DuanXT12} propose the heterogeneous feature augmentation method  which augments the homogeneous common features $\Tilde{\bx}$ by using a SVM-style approach with the help of the labels. The domain loss is defined to be the hinge loss $L(\{\mathcal{D}_S\};\Theta_S) = \frac{1}{n_S}\sum_i^{n_S} \max \left(0, 1-y_{S_i}(\bw_S^\top \Tilde{\bx}_{S_i}-b) \right)$, $L(\{\mathcal{D}_T\};\Theta_T) = \frac{1}{n_T}\sum_i^{n_T} \max \left(0, 1-y_{T_i}(\bw_T^\top \Tilde{\bx}_{T_i}-b) \right)$. %In \cite{ijcai2017-454}, cross-entropy loss $\L() = \log \left( \sum_k^K \exp(\bw_k^\prime\Tilde{\bx})\right)$ is deployed, where $\bw_k$ is the parameter vector for the $k$-th class.  

\subsection{Low-Resource Sequence Labeling}
Sequence labeling tasks such as named entity
recognition (NER) and part-of-speech (POS) tagging are fundamental problems in natural language processing (NLP). In recent, a variety of deep models have been proposed for sequence labeling, which generalize well on the new entities by automatically learning features from the data. However, when the annotated corpora is small, especially in the low resource scenario, the performance of these methods degrades significantly
because the hidden representations cannot be
learned adequately.
To enable knowledge transfer for low-resource sequence labeling (LRSL). The recent methods like \cite{fang2017model} are proposed based on cross-resource word embedding, which bridge the low- and high-resources and enable knowledge transfer. 
% The first group  \cite{chen2010onjointly} primarily assumed a large parallel corpus and focused on exploiting them to project information from high- to low-resource. Unfortunately, such a large parallel corpus may not be available for many low-resource languages. More recently, the second group of methods that is based on cross-resource word embedding \cite{fang2017model} was proposed to bridge the low- and high-resources and enable knowledge transfer. 

To implement the domain loss $L(\cdot)$, most existing works of sequence labeling usually use a linear chain model based on the first-order Markov chain structure, termed the chain conditional random field (CRF). For example, \citeauthor{zhou2019roseq} employ a CRF as the label decoder to induce  a  probability  distribution over the label sequences that is conditioned on the word-level latent features $\bh_t$. In the decoder, there are two kinds of cliques: local cliques and transition cliques. 
Specifically, local cliques correspond to the individual elements in the sequence, and transition cliques reflect the evolution of states between two nearby elements at time $t-1$ and $t$. The transition distribution is defined as $\theta$, and thus a linear-chain CRF can be formally written as $p(\mathbf{y}|\mathbf{\bh}_{1:T})  = \frac{1}{Z(\mathbf{\bh}^m_{1:T})}\exp{\left\{\sum_{t=2}^{T} \theta^m_{y_{t-1},y_{t}} + \sum_{t=1}^{T}\bW^m_{y_t}\bh^m_t\right\}}
$
where $Z(\mathbf{\bh}_{1:T})$ is a normalization term and $\mathbf{y}=y_{1:T}$ is the sequence of predicted labels. 

The model is optimized through maximizing this conditional log likelihood which acts as the following domain discriminative loss,
\begin{eqnarray}
\small
  L(\{\mathcal{D}_S,\mathcal{D}_T\}) = -\sum_{\mathcal{D}_S}\log (p(\mathbf{y}|\mathbf{\bh}^S_{1:T})) - \sum_{\mathcal{D}_T}\log (p(\mathbf{y}|\mathbf{\bh}^T_{1:T})), \nonumber
  \label{eq-crf-standard}
\end{eqnarray}
where $\bh_t^S,\bh_t^T$ are the word-level representations for the $t$-th input word of a sequence from source domain and the target domain, respectively. 

% $
% \ell_S = - \sum_{i} \log p(\mathbf{y}|\bh_{1:T}),~
% \ell_T = - \sum_{i} \log p(\mathbf{y}|\bh_{1:T}).
% \label{eq-s-crf}
% $

To model the domain discrepancy loss $M(\cdot)$, adversarial discriminator is introduced with a great success in low-resource sequence labeling tasks. For example, 
%\citeauthor{gui2017part} apply the adversarial discriminator to POS tagging for Twitter. \citeauthor{kim2017cross} proposed a language discriminator to enable language-adversarial training for cross-language POS tagging. 
\citeauthor{8778733} propose the generalized resource-adversarial discriminator (GRAD) as a domain discrepancy loss to enable adaptive weights for each sample.  As a result, the imbalance issue in training size between two domains could be largely alleviated. Most existing adversarial discriminators could be understood in GRAD with the following formulation, 
\begin{equation}
\begin{split}
M (\{\mathcal{D}_S,\mathcal{D}_T\}) =&-\sum_{i} \{ \bI_{i\in \mathcal{D}_S}\alpha(1-{r}_i)^\gamma \log {r}_i \\
&+\bI_{i\in \mathcal{D}_T}(1-\alpha){r}_i^\gamma\log (1-{r}_i)\} \nonumber
\end{split}
\label{eqn:GRAD}
\end{equation}
where $\bI_{i\in \mathcal{D}_S}, \bI_{i\in \mathcal{D}_T}$ are the identity functions which denote whether a sentence is from high resource (source) and low resource (target). $\alpha$ is a weighting factor to balance the loss contribution from the high and low resource. The parameter $(1-{r}_i)^\gamma$  (or $ {r}_i^\gamma$)  controls the loss contribution from individual samples by measuring the discrepancy between prediction and true label (easy samples have smaller contribution).% In other words, $\gamma$ scales the contrast of the loss contribution between the hard and easy samples. 

\section{Learning Using Privileged Information}
Different from the heterogeneous transfer learning where  either the unlabeled cross-domain correspondences or labels for two domains are given, Learning Using Privileged Information (LUPI) is a novel setting proposed by \citeauthor{vapnik2009new}. In brief, LUPI assumes that the triplets $\{\bx_i,\bx_i^*,y_i\}_{i=1}^N$ ($\bx_i \in \mathcal{D}_1, \bx_i^* \in \mathcal{D}_2$) are available in training phrase  but only  $\bx$ is available in the testing phrase, i.e., the testing phrase is without access to privileged information (PI) $\bx^*$.  Such a setting has been  extended into the application scenarios such as image classification \cite{lambert2018deep}, information retrieval \cite{zhou2016transfer}, and so on. Recently, \citeauthor{LopSchBotVap16} unifies the privileged information and distillation into a new framework called generalized distillation.  In mathematics, the objective functions of these studies share the following formulation,
 \begin{equation}
%\begin{split}
\mathcal{J} =\underbrace{L(\mathcal{D}_1)}_{\text{Task Loss}}+\lambda_1\underbrace{ M(\{L(\mathcal{D}_1),\mathcal{D}_2\};\Theta_M)}_{\text{PI regularization loss}} + \lambda_2\underbrace{R(\mathcal{D}_1,\mathcal{D}_2)}_{\text{Regularizer}}\nonumber
%\end{split}
\end{equation}
where $L(\mathcal{D}_1)$ is the task loss. The PI regularization loss is defined on $L(\mathcal{D}_1)$ and the privileged information $\mathcal{D}_2$. Different from the  heterogeneous transfer learning or multi-view learning which directly uses the cross-domain correspondences as input to minimize the cross-domain loss $M(\cdot)$, LUPI aims at leveraging the privileged information to model the task loss to achieve better generalization.

\subsection{Multi-Object Recognition with PI}
Multi-object recognition is usually recast into a  Multi-instance multi-label (MIML) learning problem. To be specific, MIML assumes there are $N$ bags in the training data, denoted by $\{X_i,Y_i\}_{i=1}^N$, where each bag $X_i$ has $m_i$ instances with the instance-level label vectors $\{\bx_{i,j},\by_{i,j}\}_{j=1}^{m_i}$ and $Y_i$ contains the labels associated with $X_i$.
In multi-object recognition, additional information such as
bounding-boxes, image captions and descriptions is often available during training phrase, which is referred as privileged information (PI).  Based on the above observation, 
\citeauthor{yang2017miml} propose using the bounding box and image caption as PI to improve multi-object recognition (MORPI) performance. Formally, for each training bag, there exists a privileged bag $X_i^*$ that contains $m_i^*$ instances $\{\bx^*_{i,j}\}_{j=1}^{m^*_i}$.
The task loss refers to the MIML-FCN loss which is defined as follows,
\begin{eqnarray}
  L(Y, f(X)) = L(\max_j(\by_j),\max_{\bx \in X}\phi(\bx)\bW) \nonumber
\end{eqnarray}
where the relation between instance-level labels $\by_j$ 
and bag-level labels $Y$ is expressed as $Y = \max_j(\by)$. The bag-level label prediction is given by $f(\bx) = \max_{\bx}\phi(\bx)\bW$. $\phi(\bx)$ is generated from a fully convolutional network, and $L(\cdot)$ can be the square loss or ranking loss for multi-object recognition. 

To utilize the privileged information, the PI regularization loss is modeled as follows,
\begin{eqnarray}
  M(L(X), X^*) = \|L(Y, f(X))-f^*(X^*)\|_2^2, \nonumber
\end{eqnarray}
where $f^*(X^*)$ denotes the output of network $f^*(\cdot)$ for an input privileged bag $X^*$. 

% In \cite{xu2015distance}, metric learning is proposed in the framework of LUPI with applications of the face verification and person-reid. In \cite{xu2015distance}, $\bbe$ is defined to be the learned distance, which is is large for the
% dissimilar pairs of samples and small for the similar pairs
% of samples and $M(\cdot)$ is implemented with  LogDet divergence. 
\subsection{Hashing with PI}
Learning to hash aims to learn a binary
code consisting of a sequence of bits from a specific dataset so that the nearest neighbor
search result in the hash coding space is as close as possible to the search result in the original space Most existing learning to hash methods assume that
there are sufficient data, either labeled or unlabeled,
on the domain of interest (i.e., the target domain)
for training. However, this assumption cannot be satisfied in some applications. To solve this problem,  \citeauthor{zhou2016transfer,DBLP:journals/tnn/ZhouZPFQG18} apply LUPI  into learning to hash, termed HPI. In these works, the task loss takes the quantization error with the following definition,
\begin{eqnarray}
  L(\mathcal{D}_1) = \|\bB^*-\bX^*\bR^*\|, \nonumber
\end{eqnarray}
 where   $\bR^*$ is a orthogonal projection matrix and $\bB^*$ is the binary code matrix for privileged data matrix $\bX^*$.
\begin{eqnarray}
M(L(\mathcal{D}_1), \mathcal{D}_2) = \|\bB^*-\bX^*\bR^*-f^*(\bX^*)\|_2^2, \nonumber
\end{eqnarray}
where $f^*$ is defined as a linear projection matrix in \cite{zhou2016transfer} and a nonlinear neural network in \cite{DBLP:journals/tnn/ZhouZPFQG18}.

There are additional regularizers to further constrain hash coding. For example, \citeauthor{zhou2016transfer} incorporate the graph structure into the formulation, namely $R(\cdot) = tr({\bB^*}^\top \bL_C \bB^*)$, where $\bL_C$ is the pre-computed graph Laplacian matrix on the target domain. 

\section{Heterogeneous Multi-task Learning}
% A general heterogeneous MTL is considered to consist of  consists of tasks with different types including supervised learning, unsupervised learning, semisupervised learning, etc \cite{zhang2017survey,yang2009heterogeneous}. The opposite to the heterogeneous MTL
% is the homogeneous MTL which consist of tasks with only one
% type. In a word, the homogeneous and heterogeneous MTL differ
% in the type of learning tasks. 
% The Heterogeneous Multi-task Learning usually can be mathematically expressed as follows,
%  \begin{equation}
% %\begin{split}
% \mathcal{J} =\underbrace{\sum_{t,i}L(f_t\{\bx_{t,i}\})}_{\text{Task Loss}}+\underbrace{\sum_{t} M(\{f_t(\bx_t)\})}_{\text{Tasks Relationship  loss}} + \underbrace{R()}_{\text{Regularizer}}
% %\end{split}
% \end{equation}
Multi-Task Learning (MTL) is a learning paradigm in machine learning, which aims to leverage the  information contained in multiple related tasks so that the generalization performance of all the tasks is improved. Different from the traditional MTL with tasks of only one type,
the heterogeneous multi-task learning (HMTL)  consists of different types of tasks including supervised learning, unsupervised learning,  reinforcement learning, and so on~\cite{zhang2017survey}. Formally, HMTL usually could be formulated as a minimization problem as below:
\begin{equation}
\begin{split}
&\min\underbrace{L(\{f_t(\mathcal{D}_{t}); \Theta_t\}_{t=1}^T)}_{\text{Tasks Loss}} + \lambda_1\underbrace{M(\{f_t(\mathcal{D}_{t}; \Theta_t)\}_{t=1}^T)}_{\text{Tasks Relationship Loss}}+ \lambda_2 \underbrace{R(\{\Theta_t\}_{t=1}^T)
}_{\text{Regularizer}}\\
&\text{s.t.} ~~\Theta^{(t)} \in \Omega_t, \Omega_t \bigcap \left(\bigcup_{i=1, i\neq t}^T\Omega_i \right) \neq \emptyset, t = 1, 2, \dots, T,\nonumber
\end{split}
\end{equation}
where $\mathcal{D}_{t}$ denotes the dataset for the $t$-th task and $\Theta_t$ is the weight parameter set for the corresponding learner with $f_t(\cdot)$. $M(\cdot)$ could be explicitly formulated to further capture the relationship across tasks. Let $\Theta_t \in \Omega_t$ denote the weight parameters, the constraint $\Omega_t \bigcap \left(\bigcup_{i=1, i\neq t}^T\Omega_i \right) \neq \emptyset$ enforces different tasks to share the same weight parameters, and thus transferring the knowledge across tasks. Either through sharing the parameters constraints space or explicit loss could facilitate the knowledge sharing among tasks. To avoid the over-fitting of the learners, one usually introduces a regularization term $R(\{\Theta_t\}_{t=1}^T)$ in the objective function. 

To improve the sequence labeling performance, \citeauthor{zhou2019learning} consider fully annotated data, incomplete annotated data and unsupervised data in a unified framework. Specifically, the task loss of the work is written as follows, 
\begin{equation}
L = \sum_{\bx_l\in\mathcal{D}_l} L_l(\bx_l;\Theta) + \sum_{\bx_p\in \mathcal{D}_p} L_p(\bx_l;\Theta) + \sum_{\bx_u\in \mathcal{D}_u} L_u(\bx_u;\Theta,\alpha), \nonumber
\end{equation}
where $L_l(\cdot), L_p(\cdot)$ denote the fully/semi-supervised CRF loss on the fully annotated data and incomplete annotated data, respectively. $L_u(\bx_u;\Theta,\balpha)$ is realized by an autoencoder-like structure and the parameter $\balpha$ is  defined by a softmax function. In \cite{zhou2019learning}, task relationship is implicitly learned through the shared parameter $\Theta\in \Omega$, where $\Omega$ is the same space for all the tasks. $R(\Theta,\balpha) = \|\Theta\|_2 + \|\balpha\|_2$ is adopted to avoid overfitting.

To model the relationship between different tasks, different methods give different loss. For example, in our recent work \cite{zhang2019nonlinear}, the task relationship is regularized through minimizing the negative correlation as below
$
M (\cdot)  = - \sum_t (f_t - \Tilde{f})^2
$
where $f_t$ is the learned neural network for the  $t$-th task and $\Tilde{f} = \frac{1}{T}\sum_t^T f_t $. 
\citeauthor{DBLP:conf/aistats/ZhouT14,JMLR:v20:13-580} propose to learn a sparse transformation matrix between two heterogeneous domains by exploiting the commonality of multiple binary classification tasks the the formulation of 
$
 M(\cdot) =\sum_{t}^{T} \|\bw_{T}^t-\bG\bw_S^t\|^2 
$
where the linear mapping $\bG$ is to minimize the difference between the target binary classifier weight $\bw_{T}^t$ and the transformed classifier weight $\bG\bw_S^t$ over $T$ tasks. $R(\cdot)=\lambda \sum_i^{d_T}\|\bg_i\|_1,$ is to enforce the sparsity on each row of $\bG$.

\section{Discussion and Future Direction}
In this paper, we provide a unified HRL framework to understand over a dozen of learning tasks/problems across multiple areas in mathematically. We deeply analyze the shared and distinct loss terms of over 10 learning tasks which are popular in machine learning,  multi-media analysis, computer vision, data mining, and natural language processing. We believe that such a unified view would benefit the AI community in both industry and academia from literature review to future directions. In addition to those aforementioned applications, there are also other interesting applications, such as person-reid, translation, visual question answering, video caption, text2image, points matching, etc, could also be revisit in the our HRL framework. Despite the recent advances, in future research, we think there are several very important and fundamental challenges of HRL deserves more attention. 
\begin{itemize}
    \item Theoretical analysis on the heterogeneous tasks/domains. A lot of experimental studies show that incorporating heterogeneous tasks/domains could boost up performances. However, there are few theoretical analysis on the relationship between the improvement of performance and the number of training data/tasks. A quantitatively metric is also expected to measure the contribution degree from different domains on final performance .  
    \item Universal pre-train model for HRL. pre-train for single domain/modality is well studied in last decades, such as ResNets, word2vec, etc.  Recently, more and more attention has been paid on pre-trainable generic representation such as ViLBERT \cite{NIPS2019_8297} for visual-linguistic tasks  and M-BERT\cite{DBLP:conf/naacl/DevlinCLT19} for multilingual tasks. However, research on the pre-train model for other applications/tasks like audio-visual tasks, video-linguistic tasks are still on the early stage.  
    \item Heterogeneous feature generation. This paper discuss those learning tasks from the perspective of feature representation or domain matching.  However, some related tasks like image/video caption, image/text style transfer requires the model to generate  heterogeneous features for  out-of-domain data. 
\end{itemize}

\bibliographystyle{named}
\bibliography{ijcai20_short}
\end{document}